\documentclass[arxivdoc]{jmlr}

\usepackage{amsfonts}
\usepackage{float}
\usepackage{longtable}% for long tables

\usepackage{booktabs}
\usepackage{times,xcolor}

\title{An SVM Based Approach for Cardiac View Planning}

 \author{\Name{Ramasubramanian Sundararajan} \Email{ramasubramanian.sundararajan@ge.com}\\
 \Name{Hima Patel} \Email{hima.patel@ge.com}\\
 \Name{Dattesh Shanbhag} \Email{dattesh.shanbhag@ge.com}\\  
 \Name{Vivek Vaidya}  \Email{vivek.vaidya@ge.com} \\
   \addr GE Global Research\\
John F. Welch Technology Centre, 122 EPIP, Whitefield Road, Bangalore 560066, India}

\begin{document}

\maketitle

\begin{abstract}
We consider the problem of automatically prescribing oblique planes (short axis, 4 chamber and 2 chamber views) in Cardiac Magnetic Resonance Imaging (MRI). A concern with technologist-driven acquisitions of these planes is the quality and time taken for the total examination. We propose an automated solution incorporating anatomical features external to the cardiac region. The solution uses support vector machine regression models wherein complexity and feature selection are optimized using multi-objective genetic algorithms. Additionally, we examine the robustness of our approach by training our models on images with additive Rician-Gaussian mixtures at varying Signal to Noise (SNR) levels. Our approach has shown promising results, with an angular deviation of less than $15^\circ$ on $90\%$ cases across oblique planes, measured in terms of average $6$-fold cross validation performance -- this is generally within  acceptable bounds of variation as specified by clinicians.
\end{abstract}

\section{Introduction}
\label{sec:intro}

Magnetic Resonance Imaging (MRI) is a widely used technique for diagnosis of patients with cardiovascular disorders. MRI images allow the clinicians to non-invasively visualize the cardiac structure as it allows for acquisition at any desired plane. 

\begin{figure}
	\centering
		\includegraphics[width=0.48\textwidth]{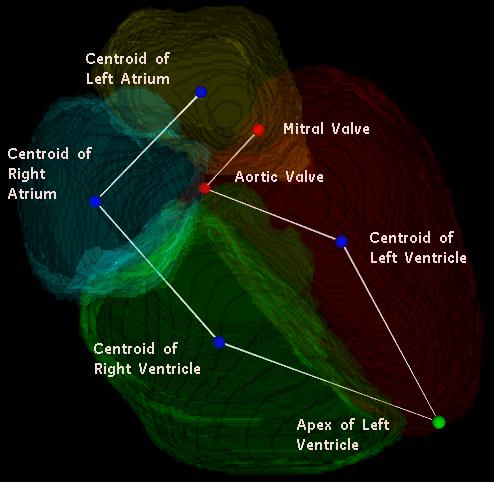}
	\caption{Cardiac landmarks and geometry}
	\label{fig:cardiac-landmarks}
\end{figure}

During conventional cardiac MRI acquisitions, the patient lies on his back in an MRI scanner (supine position) and the operator starts by performing a low-resolution axial localizer scan i.e., acquiring a series of cross sectional image slices in the cardiac region along the head-foot direction. While this allows the operator to identify the location of the heart, gaining clinically relevant information requires that slice acquisition be made along the 3D orientation of the heart. Therefore the operator has to manually estimate the orientation of the cardiac anatomy in 3D space (see Figure \ref{fig:cardiac-landmarks}) to reliably plan the views of interest for a radiologist. 

\begin{figure}
	\centering
		\includegraphics[width=0.48\textwidth]{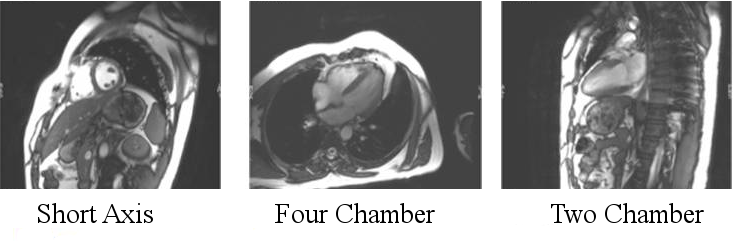}
	\caption{Cardiac cross-sectional views}
	\label{fig:heart-views}
\end{figure}

The views of clinical importance are two chamber, four chamber and short axis views, especially because they provide alternate views of the left ventricle which pumps oxygenated blood to the rest of the body. Figure \ref{fig:heart-views} illustrates an example of these oblique acquisition views for reference. 

Due to complexity of the anatomy, this is a highly operator-dependent and time consuming process. Thus, it becomes important to have a system which can automatically and reliably prescribe the planes for various views of the heart, and therefore impact both the quality and time taken for a cardiac MRI exam.

Compared to other anatomies (e.g. head, spine), cardiac MRI is more challenging due to the elastic 3D motion of the heart, affecting both MR acquisition and image quality. Additionally, the problem is complicated by variations arising due to scanner field strength, pathology, mild metal artifacts, and breath hold variation.

Existing approaches in cardiac MRI view planning mostly involve matching with a reference image with labeled thoracic anatomy (atlas-based matching) to determine the location and orientation of the left ventricle (LV). These approaches primarily rely on using cardiac-based landmarks (e.g. mitral valve) which are susceptible to cardiac motion-related variations. Approaches therein include:

\begin{itemize}
\item Fuzzy surfaces to  model thoracic anatomy \cite{Lelieveldt2001}.
\item Active shape models with anchor detection to fit a 3D mesh model of the LV, guided by pose and scale estimates arrived at using probabilistic boosting trees \cite{Lu2011}.
\item Model-based segmentation using a triangulated surface model composed of seven major parts of the heart and thoracic vessels \cite{Frick2011}.
\item Graph-based LV segmentation \cite{Darrow2008,Darrow2009}.
\item Semi-automated segmentation of LV and Right Ventricle (RV) blood pools using Gaussian Mixture Models and Expectation Maximization \cite{Jackson2004}.
\end{itemize}

We propose an atlas-free machine learning approach that uses a combination of features from graph-based segmentation \cite{Felzenszwalb2004} of the LV and adjacent anatomy to train Support Vector Machine regression models to identify the LV centroid as well as plane angulations. Our approach is less sensitive to patient-specific variations, while also improving robustness by using features that are less sensitive to cardiac motion (namely, features external to the cardiac region such as torso dimensions).

Feature selection and model complexity optimization for the SVMs are performed using a multiobjective genetic algorithm. We note that our approach yields good accuracy and robustness to variational modes such as: scanner field strength, pathology, mild metal artifacts, and breath hold variation without the need for time consuming mitigation strategies such as a backup navigator localizer \cite{Frick2011}.

It has been demonstrated in \cite{Yic2012} that training learning algorithms on deliberately noised data can enhance model generality and robustness in vision problems.  We adapt this methodology to MRI workflow automation by training our models on data with controlled SNR variation achieved through additive Rician noise. We test our proposed algorithm on both patient and volunteer data containing $13$ modes of pathology variations. The purpose of this is to prove the robustness of our approach to even the most SNR limited clinical applications, such as highly accelerated imaging. 
%Figure \ref{fig:Normal} illustrates axial scans with varying levels of noise.

The organization of this paper is as follows: Section \ref{sec:MaterialsAndMethods} describes the method proposed for localizing and calculating the angulations for the oblique planes. Section \ref{sec:Experiments} discusses the experimental results from the approach. Section \ref{sec:Conclusion} concludes and suggests directions for future work.

\section{Methodology}
\label{sec:MaterialsAndMethods}

Our experimental procedure can be summarized as:
 \begin{description}
 \item[Dataset] Obtain axial localizer scans for each patient.
 \item[Noise addition] Generate further images at varying SNR levels using additive Rician noise.
 \item[Left Ventricle Centroid identification using SVM] Use an SVM regression model (whose features and complexity parameters are optimized using a multi objective Genetic Algorithm) to localize the left ventricle of the heart.
 \item[Initial Short Axis estimation] Segment the blood pool of the left ventricle of the heart and estimate the short axis angulation.
 \item[Oblique Plane Estimation using SVM] Pass the Short Axis estimate along with other features from nearby anatomy to an SVM for generating cardiac scan geometry (SA,4CH,2CH).
 \item[Evaluation] Evaluate against the ground truth angulations prescribed by an MRI operator.
 \end{description}

 These steps are described in greater detail in the following subsections.

\subsection{Data Description}
\label{sec:DataAndPopulationStructure}

 We obtained $35$ ($19$ clinical, $16$ volunteer, comprising $6$ sub-populations) axial localizer scans acquired at $1.5T$ and $3T$ from varying MRI scanners. We selected one case from each sub-population at random as a test image, the remaining $29$ cases were used for model building. The short-axis, 4-chamber and 2-chamber views were manually prescribed by the technologists and used as ground truth for evaluation.

To ensure robustness to pathology, this dataset contained  $13$ modes of pathology variations namely: (a) Mild LV Hypertrophy, (b) Mild Aortic Regurgitation, (c) Moderate LV Dilatation with Reduced Ejection Fraction, (d) Right Ventricular Dysplasia, (e) Septal Hypertrophy, (f) Mitral Regurgitation, (g) Metal Wires in Sternum, (h) Post-surgical, (i) Pectus Excavatum (stoved in chest), (j)Hiatus Hernia,  (k) Pleural Fluid, (l) Mild LV and Septal Thickening, (m)Possible Arrythmogenic Right Ventricular Dysplasia.

\subsubsection{Noise Addittion}

In \cite{Gudbjartsson1995}, it was demonstrated that multiple coil MRI noise follows a Rician distribution in the background. Though noise foreground objects in MRI approximate a Gaussian distribution, the foreground to background transitions (such as organ boundaries) are better modeled with Rician noise instead of additive Gaussian noise. Our procedure for adding Rician noise to produce controlled  SNR variants followed \cite{constantinides2005signal,Shanbhag2008}. To simulate multiple coil Rician noise we first inject gaussian noise into the real and imaginary channels. We then perform a root sum-of-square (RSS) magnitude reconstruction using the original number of coils (taken from the DICOM data). We then add the noise matrix generated by the previous step to the original data matrix. We perform this procedure iteratively until the defined SNR limit is reached. We calculate SNR as a ratio of signal over total volume to the standard deviation of background noise for defined regions of interest. For our experiments, we generated reduced SNR images at values of $\{30, 25, 20, 15, 10\}$.

\subsection{Left Ventricle Localization}
\label{sec:LocalizeLeftventricle}

The first step in our proposed approach is to estimate the LV centroid as this serves as an initialization to short axis estimation elaborated in \ref{sec:InitialEstimationOfShortAxis}. We build a SVM regression model utilizing the features of the nearby anatomy to estimate the LV centroid. The rationale behind the approach is as follows: The LV Centroid by itself does not have strongly differentiating image characteristics such as shape, size or texture. However, there are various structures around it which differentiate themselves in the image space. We propose to use these structures to estimate the location of our point of interest.

Our feature extraction step starts with extraction of eight features. These include Torso Height, Torso Width, Left Lung Centroid (x,y,z) and the Right Lung centroid (x,y,z). Torso dimensions are obtained by first segmenting the torso through a bounding box in an axial image. 

We build Support Vector Machine (SVM) regression models for estimating the LV Centroid. The model complexity for the SVM  is determined by the Gaussian kernel spread $\gamma$ and the penalty parameter $C$ \cite{Cristianini2000,Hsu2000}. Since the number of patient images available for training is quite low, we wish to build a model that is simple and uses the most effective and smallest feature subset.  However, the best model complexity for a given feature subset may not work for a different subset; this entails a search for the best model complexity for each feature subset under consideration. We address the resultant computational complexity in the following manner: We search through a combined solution space where each solution point is defined in terms of a feature subset as well as the model complexity for that subset. We perform this search using an evolutionary multiobjective optimization algorithm, namely Non-Dominated Sorting Genetic Algorithm II (NSGA-II) \cite{Deb2002}, such that the $k$-fold cross-validation error (mean absolute deviation) is minimized, while also trying to find a parsimonious feature subset for the model.

We report the predictive performance of our models in Section \ref{sec:Experiments}, as 3D angular deviation from ground truth and average Euclidean distance between predicted and actual LV centroid locations, using $k$-fold cross-validation.

\subsection{Initial Estimation of Short Axis}
\label{sec:InitialEstimationOfShortAxis}

\begin{figure}
	\centering
		\includegraphics[width=0.3\textwidth]{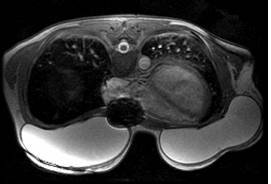}
	\caption{Example of pathology (presence of implants)}
	\label{fig:path-ex}
\end{figure}

Once the LV centroid has been estimated, the next step in our proposed approach is to calculate an initial estimate of the short axis vector of the left ventricle. The methodology utilizes the efficient-graph based segmentation described in \cite{Felzenszwalb2004}. The approach can be summarized in four steps: Localize the left ventricle, segment the blood pool, fit an ellipse to the left ventricle and compute principal components of the ellipse to arrive at a short axis estimate. The reliability of this approach declines in the presence of significant pathology ( Figure \ref{fig:path-ex} shows an example). In order to improve upon the results obtained using this technique, we use a set of machine learning models described in Section \ref{sec:FinalModelStructure}.

\subsection{SVM Based Refinements of Angulations}
\label{sec:FinalModelStructure}

After we find an initial estimate of the short axis, we arrive at our final estimates of short axis, four chamber and two chamber angulations by using SVM regression models. We first decompose each plane into its Azimuth and Elevation angles. Note that, for the two-chamber view, some operators choose to plan purely on an axial localizer resulting in a constant Elevation of  $90^{\circ}$ -- therefore, the model is built only for those cases where this is not done.

For each of the above, we build a support vector machine regression model \cite{Cristianini2000,Hsu2000} to predict the angle as a function of the patient-level features.  A total of six features are considered, including the size of the torso, body shape/size, fat estimation, the ratio of lung sizes etc. Apart from this, the initial short axis Azimuth and Elevation estimates from the method described in Section \ref{sec:InitialEstimationOfShortAxis} are also considered.

\begin{figure}
	\centering
		\includegraphics[width=0.25\textwidth]{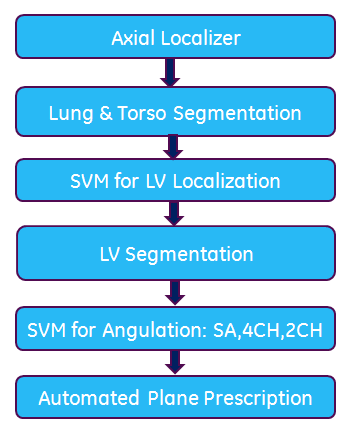}
	\caption{ Online processing stages}
	\label{fig:onlineP}
\end{figure}

We optimize the SVM regression models as well as the features used for Azimuth and Elevation for each of the oblique planes in the same way that we do for the left ventricle centroid, i.e., using a genetic algorithm. We combine the Azimuth and Elevation predictions to arrive at the predicted 3D plane orientation for each oblique plane, and measure overall performance at the plane level in terms of the 3D angle between the predicted orientation and the ground truth. We report the predictive performance of our angulation models in Section \ref{sec:Experiments}, both in terms of mean absolute deviation of Azimuth and Elevation predictions from ground truth as well as in terms of the 3D angle, for $k$-fold cross-validation.  Figure \ref{fig:onlineP}) describes the operation of the complete algorithm when applied online.

\section{Experimental Results}
\label{sec:Experiments}

We divide our data into training and testing data sets as described in Section \ref{sec:DataAndPopulationStructure}. We use a $6$-fold cross validation procedure, while ensuring that a single patient's data (original and noised images) is not distributed among different folds.  Table 1 below presents the cross validation when training with original and noised images. We see a slight improvement in overall numbers in the noised cross validation experiment in spite of the reduction in SNR (Signal to Noise Ratio).  A potential implication of being able to maintain accuracy despite a $3\times$ reduction in SNR is that by the MRI scaling relationship of

\begin{equation}\label{eq:snr}
  SNR \approx  \sqrt{Acquisition}
\end{equation}

acquisition time acceleration of up to $9\times$ can be examined.  Further online experiments will be required in order to validate this hypothesis.

Table 2 below compares the performance of the pure and noised model on 6 cases we removed prior to model construction.  We added Rician noise to the test population to create reduced SNR image subsets from 30 to 10.  Localization, SA, and 4CH results are better for the noised model, while 2CH results are better for the pure model.   The overall trend appears to support the hypothesis that noised models are more robust to low SNR data, though our 2CH result requires additional investigation.

\begin{table}[htbp]
  \centering
    \begin{tabular}{|r|r|r|r|}
    \hline
    \textbf{Model} & \textbf{Metric} & \textbf{Original} & \textbf{$SNR \geq 10$}\\
    \hline
    LV Centroid & Euclidean distance & 9.64  & 9.26 \\
    Short Axis & 3D Angle & 8.94  & 8.18 \\
    $4$ Chamber & 3D Angle & 11.53 & 10.56 \\
    $2$ Chamber & 3D Angle & 7.33  & 7.11 \\
    \hline
    \end{tabular}%
  \caption{$6$-fold CV performance of models trained on only original images and on $SNR \geq 10$ images}
  \label{tab:kfoldres}%
\end{table}%

\begin{table}[htbp]
  \centering
    \begin{tabular}{|r|r|r|r|}
    \hline
    \textbf{Model} & \textbf{Metric} & \textbf{Original} & \textbf{$SNR \geq 10$}\\
    \hline
    LV Centroid & Euclidean distance & 15.38  & 13.75 \\
    Short Axis & 3D Angle & 12.86  & 12.05 \\
    $4$ Chamber & 3D Angle & 11.95 & 11.40 \\
    $2$ Chamber & 3D Angle & 9.15  & 11.83 \\
    \hline
    \end{tabular}%
  \caption{Test sample performance on images with $SNR \geq 10$, for models trained on only original images and on $SNR \geq 10$ images}
  \label{tab:testres}%
\end{table}%

\section{Directions for future work}
\label{sec:Conclusion}

 Our proposed approach needs to be tested more extensively on volunteers and clinical populations online, in order to better gauge its effectiveness on diverse pathology modes. This includes benchmarking our algorithm against existing methods proposed in the literature. Also, operator variation on plane prescription methodologies has currently been ignored: by accounting for this, one may be able to build an adaptive model that arrives at an initial estimate using data from various sites/operators and fine-tunes it using the relevant data that illustrates a specific operator's approach. Also, since the covariates used in the model pertain to size/shape of the patient, the ability of the model to extrapolate well across people with diverse physical characteristics needs to be examined in more detail.

The method used for addition of noise is consistent with structural assumptions that can be made with regard to deterioration in image acquisition quality while maintaining patient orientation and acquisition protocol. This meant that the ground truth obtained for noise-less images was applicable even with the addition of noise, and therefore gave us a larger training sample to use for model building. However, the approach also suggests that our algorithm may be applicable for situations where noisier (albeit faster) image acquisition is made -- further study needs to be done to verify this hypothesis.

Furthermore, we feel that this general approach to handling patient and pathology variation is applicable to other challenging MRI acquisitions such as spine and liver, and recommend further exploration in these directions.

\bibliography{otc}

\begin{thebibliography}{14}
\providecommand{\natexlab}[1]{#1}
\providecommand{\url}[1]{\texttt{#1}}
\expandafter\ifx\csname urlstyle\endcsname\relax
  \providecommand{\doi}[1]{doi: #1}\else
  \providecommand{\doi}{doi: \begingroup \urlstyle{rm}\Url}\fi

\bibitem[Constantinides et~al.(2005)Constantinides, Atalar, and
  McVeigh]{constantinides2005signal}
C.D. Constantinides, E.~Atalar, and E.R. McVeigh.
\newblock Signal-to-noise measurements in magnitude images from nmr phased
  arrays.
\newblock \emph{Magnetic Resonance in Medicine}, 38\penalty0 (5):\penalty0
  852--857, 2005.

\bibitem[Cristianini and Shawe-Taylor(2000)]{Cristianini2000}
N.~Cristianini and J.~Shawe-Taylor.
\newblock \emph{An Introduction to Support Vector Machines}.
\newblock Cambridge University Press, 2000.

\bibitem[Darrow et~al.(2008)Darrow, Vaidya, Govenkar, Mullick, and
  Foo]{Darrow2008}
R.~Darrow, V.~Vaidya, A.~Govenkar, R.~Mullick, and T.K. Foo.
\newblock One touch imaging for reduced cardiac workflow.
\newblock In \emph{Proceedings of ISMRM}, 2008.

\bibitem[Darrow et~al.(2009)Darrow, Vaidya, Mullick, Govenkar, Ho, Fung,
  Varghese, and Foo]{Darrow2009}
R.~Darrow, V.~Vaidya, R.~Mullick, A.~Govenkar, V.~Ho, M.~Fung, B.~Varghese, and
  T.~Foo.
\newblock Ground truth evaluation of one touch cardiac imaging.
\newblock In \emph{Proceedings of ISMRM}, 2009.

\bibitem[Deb et~al.(2002)Deb, Pratap, Agarwal, and Meyarivan]{Deb2002}
K.~Deb, A.~Pratap, S.~Agarwal, and T.~Meyarivan.
\newblock A fast and elitist multiobjective genetic algorithm: Nsga-ii.
\newblock \emph{IEEE Transactions on Evolutionary Computation}, 6:\penalty0
  182--197, 2002.

\bibitem[Felzenszwalb and Huttenlocher(2004)]{Felzenszwalb2004}
P.~F. Felzenszwalb and D.~P. Huttenlocher.
\newblock Efficient graph-based image segmentation.
\newblock \emph{International Journal of Computer Vision}, 59, 2004.

\bibitem[Frick et~al.(2011)Frick, Paetsch, den Harder, Kouwenhoven, Heese,
  Dries, Schnackenburg, de~Kok, Gebker, Fleck, Manka, and Jahnke]{Frick2011}
M.~Frick, I.~Paetsch, C.~den Harder, M.~Kouwenhoven, H.~Heese, S.~Dries,
  B.~Schnackenburg, W.~de~Kok, R.~Gebker, E.~Fleck, R.~Manka, and C.~Jahnke.
\newblock Fully automatic geometry planning for cardiac magnetic resonance
  imaging and reproducibility of functional cardiac parameters.
\newblock \emph{Journal of Magnetic Resonance Imaging}, 34\penalty0
  (2):\penalty0 457467, 2011.

\bibitem[Gudbjartsson and Patz(1995)]{Gudbjartsson1995}
H.~Gudbjartsson and S.~Patz.
\newblock The rician distribution of noisy mri data.
\newblock \emph{Magnetic Resonance in Medicine}, 34\penalty0 (6):\penalty0
  910--914, 1995.

\bibitem[Hsu et~al.(2000)Hsu, Chang, and Lin]{Hsu2000}
C-W. Hsu, C-C. Chang, and C-J. Lin.
\newblock A practical guide to support vector classification.
\newblock 2000.

\bibitem[Jackson et~al.(2004)Jackson, Robson, Francis, and Noble]{Jackson2004}
C.~E. Jackson, M.~D. Robson, J.~M. Francis, and J.~A. Noble.
\newblock Computerised planning of the acquisition of cardiac mr images.
\newblock \emph{Computerized Medical Imaging and Graphics}, 28\penalty0
  (7):\penalty0 411--418, 2004.

\bibitem[Lelieveldt et~al.(2001)Lelieveldt, van~der Geest, Lamb, Kayser, and
  Reiber]{Lelieveldt2001}
B.~P.~F. Lelieveldt, R.~J. van~der Geest, H.~J. Lamb, H.~W.~M. Kayser, and
  J.~H.~C. Reiber.
\newblock Automated observer-independent acquisition of cardiac short-axis mr
  images: A pilot study.
\newblock \emph{Radiology}, 221\penalty0 (2):\penalty0 537--542, 2001.

\bibitem[Lu et~al.(2011)Lu, Jolly, Georgescu, Haye, Speier, Schmidt, Bi,
  Kroeker, Comaniciu, Kellman, Mueller, and J.]{Lu2011}
X.~Lu, M.P. Jolly, B.~Georgescu, C.~Haye, P.~Speier, M.~Schmidt, X.~Bi,
  R.~Kroeker, D.~Comaniciu, P.~Kellman, E.~Mueller, and Guehring J.
\newblock Automatic view planning for cardiac mri acquisition.
\newblock In \emph{Proceedings of MICCAI}, volume 6893, pages 479--486, 2011.

\bibitem[Shanbhag et~al.(2008)Shanbhag, Narayanan, Krishnan, Hervo, and
  Mullick]{Shanbhag2008}
D.~D. Shanbhag, A.~Narayanan, K.~Krishnan, P.~Hervo, and R.~Mullick.
\newblock Snr performance of automated geodesic active contour based liver
  segmentation.
\newblock In \emph{ISMRM}, 2008.

\bibitem[Yichuan et~al.(2012)Yichuan, Salakhutdinov, and Hinton]{Yic2012}
T.~Yichuan, R.~Salakhutdinov, and G.~Hinton.
\newblock Robust boltzmann machines for recognition and denoising.
\newblock \emph{IEEE Conference on Computer Vision and Pattern Recognition},
  pages 2264 -- 2271, 2012.

\end{thebibliography}

\end{document}